%% file: bare_jrnl.tex
\documentclass[journal]{IEEEtran}
\ifCLASSINFOpdf
\else
\fi
\usepackage{soul}
\usepackage{url}
\usepackage{latexsym}
\usepackage{ulem}
\usepackage{booktabs}
\usepackage{amsfonts}
\usepackage{multicol}
\usepackage{multirow}
\usepackage{hyperref}
\usepackage{graphicx}
\usepackage{amsmath}
\usepackage{booktabs}
\usepackage{colortbl}
\usepackage{stfloats}
\usepackage{comment}
\usepackage{makecell}
\usepackage[noend]{algpseudocode}
\usepackage{algorithmicx,algorithm}
\usepackage{pifont}
\usepackage{color}
\usepackage{microtype}
\usepackage{cleveref}
\newcommand{\tabincell}[2]{\begin{tabular}{@{}#1@{}}#2\end{tabular}}
\usepackage{cite}

\begin{document}
\title{Understand me, if you refer to Aspect Knowledge: 
Knowledge-aware Gated Recurrent Memory Network}

\author{Bowen~Xing and Ivor W. Tsang,~\IEEEmembership{Fellow,~IEEE} % <-this % stops a space
\thanks{Bowen Xing is with Australian Artificial Intelligence Institute (AAII), University of Technology Sydney.
Ivor W. Tsang is with the A$^*$STAR Centre for Frontier AI Research (CFAR), and also with Australian Artificial Intelligence Institute (AAII), University of Technology Sydney. \qquad \qquad \qquad\qquad \qquad\qquad \qquad 
E-mail: Bowen.Xing@student.uts.edu.au, ivor\_tsang@ihpc.a-star.edu.sg}
\thanks{© 2022 IEEE.  Personal use of this material is permitted.  Permission from IEEE must be obtained for all other uses, in any current or future media, including reprinting/republishing this material for advertising or promotional purposes, creating new collective works, for resale or redistribution to servers or lists, or reuse of any copyrighted component of this work in other works.}
}% <-this % stops a space


\markboth{Journal of \LaTeX\ Class Files,~Vol.~14, No.~8, August~2015}%
{Shell \MakeLowercase{\textit{et al.}}: Bare Demo of IEEEtran.cls for IEEE Journals}



\maketitle

\input{abstract}
%And empirical study demonstrates the advantages of our model.
%The ablation study empirically verifies the efficacies of different components and comprehensive analysis demonstrates that the aspect knowledge is consistently beneficial to ASC task and the necessity of the recurrent manner of the knowledge integration in KaGR-MN.


% \fi
\IEEEpeerreviewmaketitle

\input{introduction}
\input{relatedwork}
%\input{Overview}
\input{method}
\input{experiment}

\input{conclusion}

\section*{Acknowledgements}
This work was supported by Australian Research Council  Grant (DP180100106 and DP200101328).
Ivor W. Tsang was also supported by A$^*$STAR Centre for Frontier AI Research (CFAR).
%\section*{Acknowledgment}


\ifCLASSOPTIONcaptionsoff
  \newpage
\fi


\normalem
\bibliographystyle{IEEEtran}
\bibliography{ref_normal}

\begin{IEEEbiography}[{\includegraphics[width=1in,height=1.25in,clip,keepaspectratio]{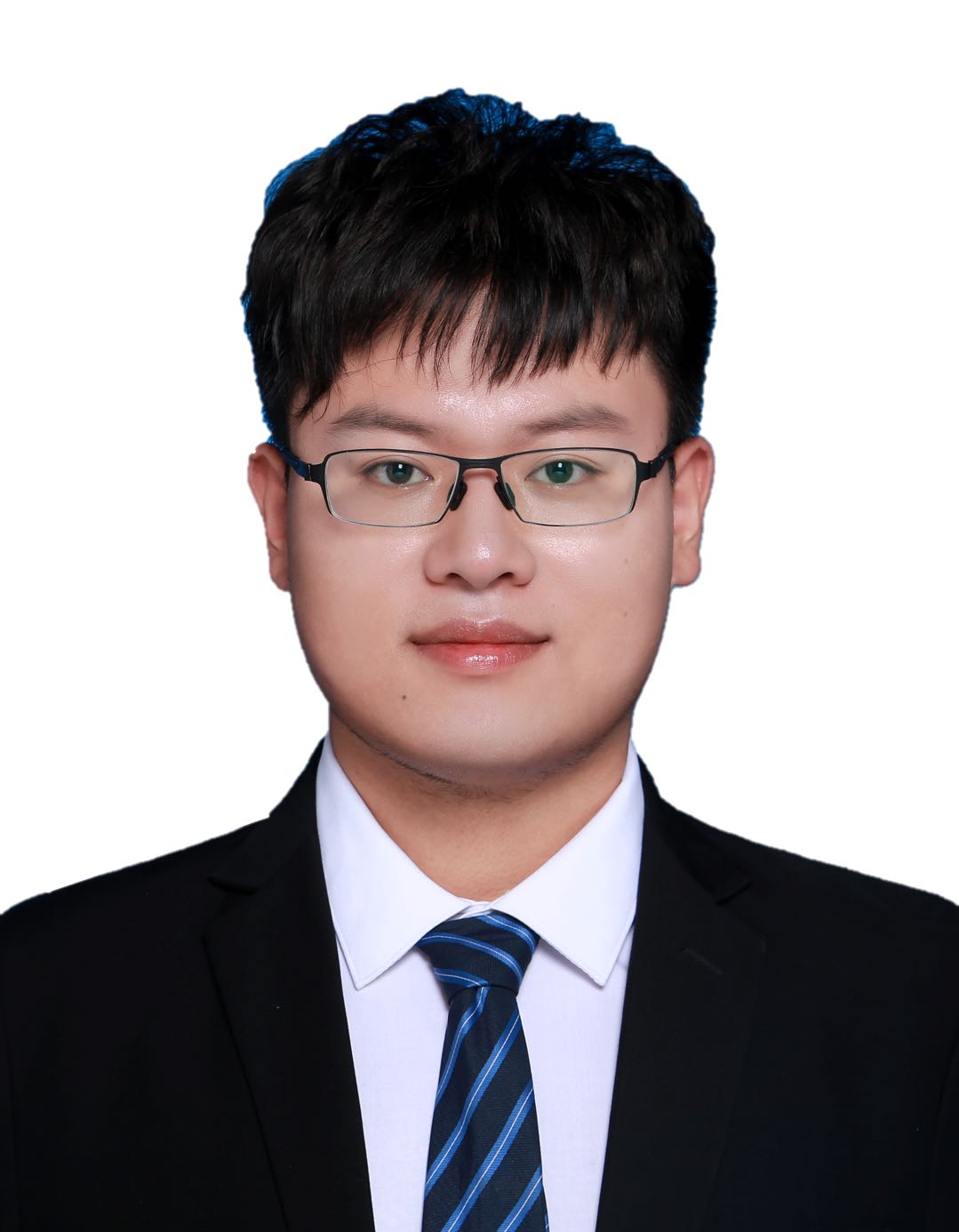}}]{Bowen Xing}
received his B.E. degree and Master degree from Beijing Institute of Technology, Beijing, China, in 2017 and 2020, respectively. He is currently a second year Ph.D student at Australian AI Institute, University of Technology Sydney (UTS).
His research focuses on graph neural network, multi-task learning, sentiment analysis and dialog system.
\end{IEEEbiography}

\begin{IEEEbiography}[{\includegraphics[width=1in,height=1.25in,clip,keepaspectratio]{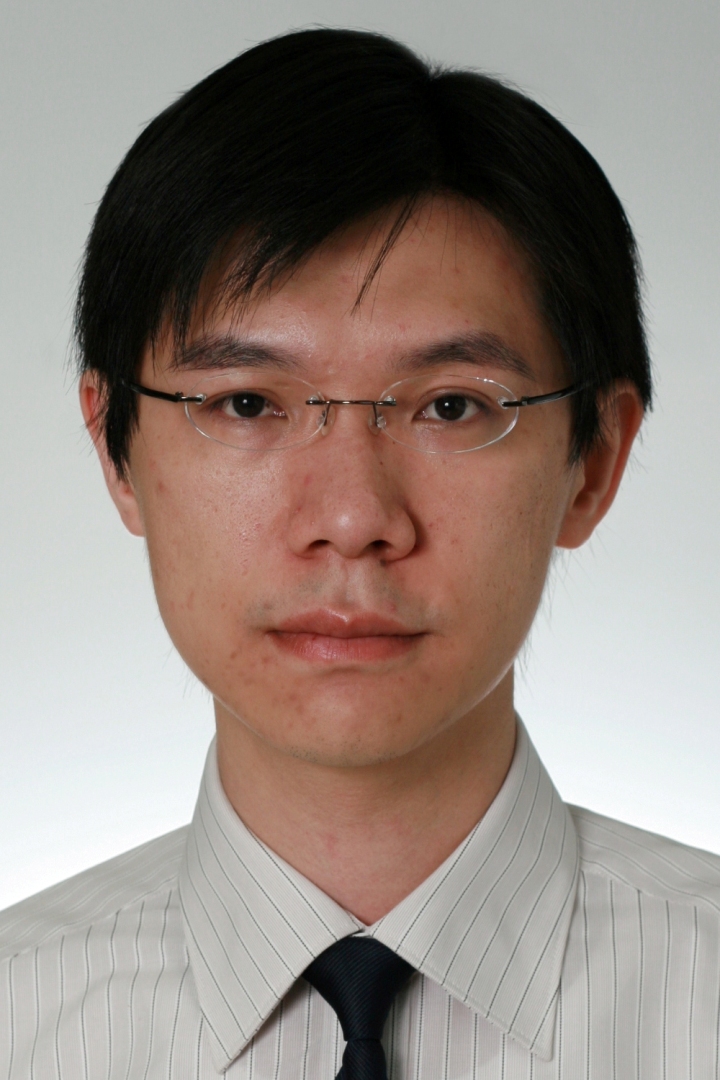}}]{Ivor W. Tsang} is an IEEE Fellow and the Director of A*STAR Centre for Frontier AI Research (CFAR). Previously, he was a Professor of Artificial Intelligence, at University of Technology Sydney (UTS), and Research Director of the Australian Artificial Intelligence Institute (AAII).
%, the largest AI institute in Australia, which is the key player to drive the University of Technology Sydney to rank 10th globally and 1st in Australia for AI research, in the latest AI Research Index. Prof Tsang is working at the forefront of big data analytics and Artificial Intelligence. 
His research focuses on transfer learning, deep generative models, learning with weakly supervision, big data analytics for data with extremely high dimensions in features, samples and labels. His work is recognised internationally for its outstanding contributions to those fields.
In 2013, Prof Tsang received his ARC Future Fellowship for his outstanding research on big data analytics and large-scale machine learning. 
In 2019, his JMLR paper ``Towards ultrahigh dimensional feature selection for big data'' received the International Consortium of Chinese Mathematicians Best Paper Award. In 2020, he was recognized as the AI 2000 AAAI/IJCAI Most Influential Scholar in Australia for his outstanding contributions to the field, between 2009 and 2019. His research on transfer learning was awarded the Best Student Paper Award at CVPR 2010 and the 2014 IEEE TMM Prize Paper Award. In addition, he received the IEEE TNN Outstanding 2004 Paper Award in 2007 for his innovative work on solving the inverse problem of non-linear representations. Recently, Prof Tsang was conferred the IEEE Fellow for his outstanding contributions to large-scale machine learning and transfer learning.
Prof Tsang serves as the Editorial Board for the JMLR, MLJ, JAIR, IEEE TPAMI, IEEE TAI, IEEE TBD, and IEEE TETCI. He serves as a Senior Area Chair/Area Chair for NeurIPS, ICML, AAAI and IJCAI, and the steering committee of ACML.
\end{IEEEbiography}

\end{document}

%% file: abstract.tex
% !TEX root = bare_jrnl.tex
\begin{abstract}
Aspect-level sentiment classification (ASC) aims to predict the fine-grained sentiment polarity towards a given aspect mentioned in a review.
Despite recent advances in ASC, enabling machines to preciously infer aspect sentiments is still challenging.
This paper tackles two challenges in ASC: 
(1) due to lack of aspect knowledge, aspect representation derived in prior works is inadequate to  represent aspect's exact meaning and property information; 
(2) prior works only capture either local syntactic information or global relational information, thus missing either one of them leads to insufficient syntactic information.
To tackle these challenges, we propose a novel
 ASC model which not only end-to-end embeds and leverages aspect knowledge but also marries the two kinds of syntactic information and lets them compensate for each other.
Our model includes four key components: 
(1) a knowledge-aware gated recurrent memory network recurrently integrates dynamically summarized aspect knowledge;
 (2) a dual syntax graph network combines both kinds of syntactic information to comprehensively capture sufficient syntactic information;
 (3)  a knowledge integrating gate re-enhances the final representation with further needed aspect knowledge;
 (4) an aspect-to-context attention mechanism aggregates the aspect-related semantics from all hidden states into the final representation.
Experimental results on several benchmark datasets demonstrate the effectiveness of our model, which overpass previous state-of-the-art models by large margins in terms of both Accuracy and Macro-F1.
To facilitate further research in the community, we have released our source code at \url{https://github.com/XingBowen714/KaGRMN-DSG\_ABSA}.
\end{abstract}

%% file: introduction.tex
% !TEX root = bare_jrnl.tex
\section{Introduction}\label{sec:introduction}
%Aspect-level sentiment classification (ASC), whose objective is to predict the fine-grained sentiments of a given aspect mentioned in a review, has received increasing attention in the sentiment analysis community. %\cite{Semeval2014,asgcn,RGAT}.
\IEEEPARstart{A}spect-level sentiment classification (ASC) \cite{tkdesurvey} is a fine-grained task of sentiment classification  or emotion recognition \cite{survysc,tetciemotionhci,tetciemotionspeech,ICDM16emotion}.
ASC aims to infer the fine-grained sentiment of a given aspect mentioned in a review.
Generally, an aspect is a noun phrase included in a review sentence. 
For example, in a review ``It took so long to get the check, while the dinner is great.'', there are two aspects (\textit{check} and \textit{dinner}) of opposite sentiments.
ASC has received increasing attention and interest from both academia and the industry due to its wide applications in real-life scenarios such as dialog systems \cite{dialogabsa}, online reviews \cite{semeval2014} and social networks \cite{twitter}.
\begin{comment}
ASC is different from the common sentence- or document-level sentiment classification due to two main characteristics.
The first one is that the sentiment unit of ASC is aspect rather than whole sentence or document, so aspect-context interaction is demanded in ASC to extract the aspect-oriented sentiment features.
The second one is that generally multiple aspects are existing in a single review and the intertwining sentiment features from different aspects make this task much more challenging.
The aspect representation and its conveyed semantics guide the aspect-context interaction to focus on aspect-related words, whose features are aggregated for sentiment classification.
Consequently, there are two crucial factors in ASC: aspect representation and aspect-context interaction.
\end{comment}
%For example, in Fig.\ref{fig: mh}, review 1 expresses a positive sentiment on  \textit{Mountain Lion OS}.
%ASC can provide more complete and in-depth results for further structured sentiment collection and analysis.

%The basis architecture of ASC models is attention-based\cite{IAN}.
%After obtaining the hidden states of the aspect and context words, the attention mechanism is adopted to identify the potential important words in review for predicting the aspect sentiment.
Prior works have noticed the importance of aspect-context interaction.
Different kinds of attention mechanisms \cite{IAN,fanmulti,songle2019} are proposed to extract aspect-relevant semantics from the hidden states of context words.
%However, attention mechanisms are not compatible to identify the related context words which are far from the aspect in the context.
And more recently, syntactic information is widely leveraged to facilitate the interactions between the aspect and its related words that are distant in context sequence. Graph Convolutional Networks (GCN) \cite{asgcn,DGEDT} and Graph Attention Networks (GAT) \cite{sagat,RGAT} are adopted to encode the syntax graphs predicted by off-the-shelf dependency parsers.
%Besides, graph neural networks such as are adopted to encode the syntactic information conveyed by the syntax graph of the review context.
\cite{asgcn,CDT} employed GCNs to capture the local syntactic information.
\cite{RGAT} proposed relational multi-head attention (Relational MHA) to capture the global relational information between aspect and each context word. % via operating on the star-shaped aspect-oriented syntax graph.
%via directly connecting the aspect and each context word with defined relations.

%Some other methods aim to improve context modeling, such as aspect-aware encoder \cite{AA-LSTM}, multi-task shared encoder \cite{hrd,transcap}, and BERT \cite{bert} encoder.
%Another line of prior work is to enhance the context modeling.
%\cite{AA-LSTM} proposed an aspect-aware encoder based on LSTM.
%\cite{hrd,transcap} leveraged the external corpus from document-level sentiment classification to enhance the encoder via multi-task learning.
%And BERT \cite{bert,aen-bert,RGAT} has been adopt as context encoder to generate better hidden states than LSTM.
\begin{comment}
\begin{figure}[t]
 \centering
 \includegraphics[width = 0.48\textwidth]{mh.pdf}
 \caption{Comparison of machines and humans on two ASC cases. Aspects are in \textcolor{gray}{gray}. Machines give incorrect predictions due to inadequate understanding of the aspects, as a result of lacking aspect knowledge.}
 \label{fig: mh}
\end{figure}
\end{comment}

However, little attention has been spent on aspect representation and its conveyed semantics.
%As they are the primary of aspect-context interaction, are they effective and informative enough to contain the exact meaning and property of the aspect?
Aspect representation and its semantics not only guide the aspect-context interaction but also provide important clues for ASC.
Despite its importance, in previous works \cite{Tencent,asgcn,RGAT}, aspect representation is simply derived by pooling the hidden states of aspect words.
In Sec. \ref{sec: casestudy} we empirically study the aspect representation generated by BERT \cite{bert}, and two cases are shown in Table \ref{table: entity}.
We can find that BERT cannot capture the exact meanings and property information of \textit{Mountain Lion OS} and \textit{iTune}, although it is one of the strongest language models.
Merely relying on pre-trained large language models cannot obtain sufficiently effective and informative aspect representation, making it hard for machines to address ASC.
In contrast, humans can easily handle ASC and we conjecture the key to master this task is to leverage the adequate aspect knowledge they often refer to as the clue.
Thinking of and leveraging the aspect knowledge are instinctive reactions of humans when they read an aspect in a review.
For example, there is a review ``Just a not bad restaurant, because the cheese and chips are both very soft.''
With the knowledge of `cheese' and `chips', humans are aware that the former should be soft and the latter should be click (not soft).
Hence it is easy for humans to infer the positive sentiment of `cheese' and the negative sentiment of `chips'.
However, in contrast, in ASC models there is no such mechanism, and aspect knowledge has not been explored or leveraged.
Inheriting this deficiency, the aspect representation and semantics derived by prior models may lose important aspect information, which hinders aspect sentiment reasoning and make ASC challenging for machines.
% In this paper, we argue that it would be beneficial to explicitly provide the aspect knowledge to ASC models.
%Table \ref{table: entity} shows the semantics of two aspects 
%Despite the advances achieved by recent methods, enabling machines to preciously infer aspect sentiments is still challenging.

%We attribute this to the fact that humans often rely on the aspect knowledge, which is hard for machines to obtain from the review context.

%In Fig.\ref{fig: mh}, we humans can easily infer the aspect sentiments based on our abundant aspect knowledge, while machines fail this task due to the lack of aspect knowledge.
%As shown in Table \ref{table: entity},  
On the other hand, both GCN and Relational MHA are useful for modeling distinct syntax graphs, but they have respective shortages:
GCN is hard to capture the global relations between aspect and its non-adjacent context words on the syntax graph;
Relational MHA fails to capture the local syntactic information among context words because they are isolated from each other on the star-shaped aspect-oriented syntax graph. 
%Both two kinds of syntactic information are beneficial. 
However,  prior works only consider one of them, resulting in insufficient syntactic information.

To tackle the aforementioned two challenges, we suggest that (1) aspect knowledge should be explicitly leveraged in ASC models; (2) both kinds of syntactic information should be combined to capture sufficient syntactic information.
We observe that there is plenty of entity descriptions in popular and easily accessible knowledge bases, such as DBpedia\footnote{https://wiki.dbpedia.org/} and Wikipedia\footnote{https://www.wikipedia.org/}.
From their statistics, there are about 6.6 Million and 50 Million entities in current DBpedia and Wikipedia datasets.
These descriptions can sufficiently represent the entities’ meanings and conveying a wealth of entities' knowledge.
  %which is easily accessible for many domain researches.
In ASC, aspects are always entities, making it more convenient to retrieve their descriptions.
%it convenient to retrieve aspects' descriptions from DBpedia and Wikipedia.

In this work, we propose a \textbf{K}nowledge-\textbf{a}ware \textbf{G}ated \textbf{R}ecurrent \textbf{M}emory \textbf{N}etwork with \textbf{D}ual \textbf{S}yntax \textbf{G}raph Modeling (\textbf{KaGRMN-DSG}) model as our solution to the two challenges.
Specifically, its novelty lies in three core modules.
The \textbf{first} one is Knowledge-aware Gated Recurrent Memory Network (KaGR-MN) which recurrently integrates the aspect knowledge into aspect representation and then context memories.
An aspect-to-description attention mechanism is devised to dynamically summarize the needed aspect knowledge from the aspect description regarding the current semantic state.
An adaptive knowledge integrating gate is designed to adaptively integrate the summarized knowledge into aspect representation.
Then a self multi-head attention is employed to contextualize the integrated knowledge and update the context memory bank.
The \textbf{second} one is Dual Syntax Graph Network (DSG-Net), which
marries the proposed Position-aware GCN and Relational MHA, then learns the dual syntactic interaction to comprehensively capture sufficient syntactic information.
%models global relational information.
%Then the mutual interaction between both kinds of syntactic information is learned to capture sufficient syntactic information. 
%which simultaneously models the local syntactic information and global relational information, and then learns their mutual interaction to make them compensate for each other, capturing sufficient syntactic information. 
%DSG-Net combines GCN and Relational MHA to let them compensate for each other’s shortage.\\
The \textbf{third} one is the knowledge integrating gate (KI Gate) which re-enhances the final representation with further needed knowledge.

We highlight our contributions as follows:\\
(1) Based on plenty of informative entity descriptions from easily accessible knowledge bases, we 
%propose a KaGRMN-DSG model to 
end-to-end embed and leverage the aspect knowledge to address ASC.\\
(2) We propose a novel KaGR-MN, which combines the advantages of LSTM, Transformer, and Memory Networks.
It recurrently embeds and integrates beneficial aspect knowledge into aspect representation and all context memories.\\
(3) We propose a dual syntax graph network, in which the local syntactic information and global relational information are combined to comprehensively capture sufficient syntactic information.\\
%The DSG-Net comprehensively captures both the local syntactic information and global relational information.\\
(4) We conduct extensive experiments on three benchmark datasets. Results show that our model achieves new state-of-the-art performances, significantly outperforming previous best results.
% by 1.92\%, 1.46\%, and 4.74\% in terms of Macro-F1 on Lap14, Res14, and Res15 datasets, respectively. 
Ablation study and further analysis validate the effectiveness of our model.

%% file: relatedwork.tex
% !TEX root = bare_jrnl.tex
\section{Related Works}
\label{sec: related work}
In early studies \cite{jiang-etal-2011-target,NRC}, sentiment classifiers were built by traditional machine learning algorithms which demanded labor-intensive feature engineering.
%As deep learning \cite{dl-nature,survey} develops, numbers of neural network models are proposed to tackle ASC task.
Most recently proposed ASC models are based on neural networks which can automatically learn representations.
Conventionally, neural ASC model contains an aspect encoder, a context encoder and an aspect-to-context attention mechanism \cite{ATAE,IAN,Tencent,fanmulti,songle2019}.

Different kinds of networks are adopted as the encoder.
As for LSTM,
\cite{TDLSTM} employed two separated LSTMs to encode the aspect-left and -right word sequences and then combined the two last hidden sates for classification;

\cite{hrd} proposed to leverage external document sentiment analysis corpus in a multi-task framework to enhance the context modeling of LSTM.
Besides, convolutional neural networks (CNN) and Memory Networks (MNs) are also exploited as encoders.
\cite{pcnn} introduced parameterized filters and parameterized gates into CNN to integrate aspect information for context encoding.
\cite{gcae} designed a gated CNN layer to extract the aspect-specific features from the context hidden states.
Based on standard MNs, \cite{tsmn} proposed the target-sensitive memory networks to focus on the impact of aspect semantics on the classification.
% (\cite{gcae,pcnn}), and Memory Network (\cite{DMN,tsmn}).

The attention mechanism is utilized to extract aspect-related sentiment features via assigning a weight to each context word regarding its relevance to the aspect. % to capture the potential sentiment sensitive words.
\cite{IAN} proposed an aspect-to-context attention and a context-to-aspect attention to study the interactions between the aspect and context.
\cite{songle2019} proposed an algorithm to automatically mine useful supervised information for the attention mechanism through the training process.

%Sometimes the important words are far from the aspect in context words sequence but near to it on the syntax graph.
However, attention mechanisms may hardly capture the important words which is far from the aspect in the input context.
As the development of graph neural networks \cite{gat,gcn,rgcn,wwwruiliwang},
 recent works utilize graph convolution network (GCN) \cite{gcn,asgcn,CDT,DGEDT} and graph attention network (GAT) \cite{gat,gat-absa,RGAT} to model the syntax graph for shortening the distance between the aspect and its sentiment trigger words and leveraging the syntactical information.
 \cite{asgcn} employed LSTM as encoder and exploit GCN to capture local syntactic information via encoding the syntax graph produced by off-the-shelf dependency parsers.
\cite{RGAT} proposed Relational MHA, which can capture the global dependency between the aspect and each context word via operating on the star-shaped aspect-oriented syntax graph.

To enhance the context modeling, \cite{hrd} and \cite{transcap} trained their models on both document-level sentiment classification and ASC tasks in the multi-task framework with a shared encoder.
\cite{AA-LSTM} proposed an aspect-aware LSTM which introduces aspect information into LSTM cells to generate better context hidden states in which more aspect-related information is retained and aspect-irrelevant information is discarded.
As BERT has proven its power of language modeling on hetergenuous NLP tasks, more recently proposed work \cite{RGAT,DGEDT} adopted BERT as the context encoder to obtain high-quality hidden states.

However, prior models neglect to leverage aspect knowledge, resulting in inadequate aspect semantics.
And the syntactic information they captured is insufficient.
In this paper, we propose KaGRMN-DSG to solve these two challenges.
There are two main differences from our model and previous works.
The first one is leveraging aspect knowledge, which is achieved by a novel KaGR-MN.
The other one is combining both of local syntactic information and global relational information, which is achieved by a DSG-Net.

%KaGRMN-DSG needs external input (the aspect description), on this point it
%As a solution, in this paper, we propose a KaGRMN-DSG model which can effectively leverage the aspect knowledge and comprehensively capture more sufficient syntactic information.

%Different kinds of architectures are adopted, such as long short-term memory network (LSTM) \cite{LSTM,TDLSTM,IAN}, convolutional neural networks (CNN) \cite{textcnn,pcnn,gcae}, memory networks \cite{end2endMN,DMN,tsmn} and capsule networks \cite{capsule,absacap,transcap}.

%% file: method.tex
% !TEX root = bare_jrnl.tex

\label{sec: model}
\begin{figure*}[t]
 \centering
 \includegraphics[width = 1.0\textwidth]{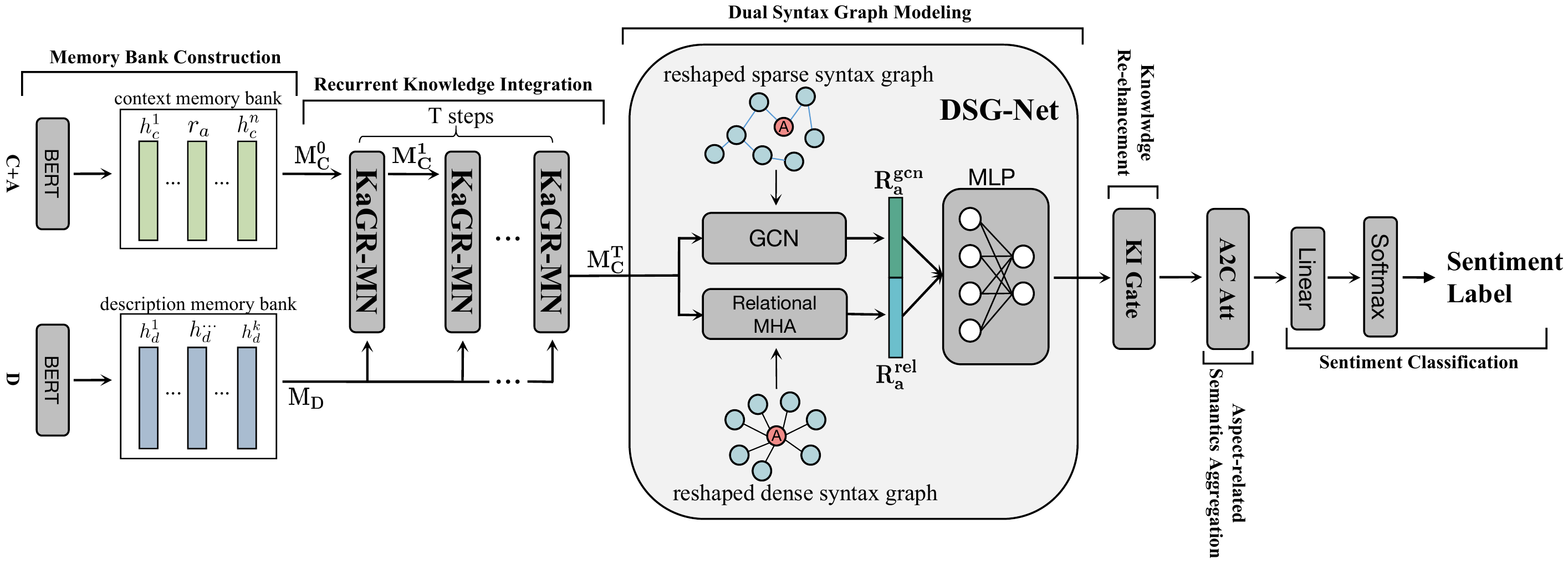}
 \caption{The architecture of KaGRMN-DSG. The internal architecture of KaGR-MN cell is shown in Fig.\ref{fig: mncell}}
 \label{fig: framework}
\end{figure*}

\section{KaGRMN-DSG}
\noindent \textbf{Overview}
The architecture of our KaGRMN-DSG model is illustrated in Fig. \ref{fig: framework}.
To extract the beneficial clues for aspects, knowledge-aware gated recurrent memory network and knowledge integration gate incorporate summarized aspect knowledge to enrich aspect representation and all context memories.
To capture sufficient syntactic information, dual syntax graph network combines local syntactic information and global relational information, then learns their mutual interaction. 
To comprehensively abstract high-level clues, aspect-to-context attention mechanism aggregates aspect-related semantics from all hidden states into the final representation.
And we believe that these modules can effective cooperate to further improve aspect sentiment reasoning.
\begin{comment}
It includes five modules of respective functions: 1) Memory Bank Construction; 2) Recurrent Knowledge Integration; 3) Dual Syntax Graph Modeling; 4) Knowledge Re-enhancement; 5) Sentiment Classification.
In this section, we introduce the details of kaGRMN-DSG following the above order.
\end{comment}

\noindent \textbf{Description Retrieval}
We use aspect ($\mathbf{A}$) to query DBpedia first and then Wikipedia to get its description ($\mathbf{D}$).
%If no description is returned, a null string is set as $\mathbf{D}$.
If multiple descriptions are returned (polysemy), the one with the highest semantic similarity to context ($\mathbf{C})$ is selected as $\mathbf{D}$.
The semantic similarity of a description candidate and review context is calculated as:
\begin{align}
avg(C)&=\frac{1}{N_C}\sum_{i=1}^{N_C}e(c_i)\\
avg(D')&=\frac{1}{N_D}\sum_{i=1}^{N_D}e(d_i)\\
sim(C,D') &= cos\big( \alpha*avg(C) + (1-\alpha)*e(dl), avg(D')\big)
\end{align}
%$\mathbf{D}$ is selected based on the semantic similarity to the context ($\mathbf{C}$). 
where $N_C$ and $N_D'$ denotes the number of words in the context and description candidate respectively, $e(w)$ denotes the word embedding\footnote{We use Glove word embedding \cite{Glove}.} of word $w$, $dl$ denotes domain label (e.g. the $dl$ of Lap14 dataset is `laptop').
Here we intuitively set $\alpha$ as 0.5 because both of the context semantics $avg(C)$ and domain information $e(dl)$ are important in selecting the correct description candidate.
The reason why we use domain label here is that sometimes there may be not enough words conveying domain-specific semantics for distinguishing the needed description.
Besides, the retrieval is enhanced with some rules, such as soft matching with lemmatization and stop word filtering.
Finally, about 70\% aspects in the datasets can be equipped with retrieved descriptions.

\subsection{Memory Bank Construction}
%BERT has proved its advantages on ASC task.
In this work, we adopt BERT to encode the description and context to produce their hidden states.
For description ($\mathbf{D}$), the formal input is $\langle\texttt{[CLS]}; \mathbf{D}; \texttt{[SEP]}\rangle$, where $\langle;\rangle$ denotes concatenation operation. The description is encoded in the single-sentence manner then a series of its hidden states is generated: $\mathbf{H_D} = \{h_d^i\in \mathbb{R}^{d_e}\}^{N_D}_{i=1}$, which is taken as the description memory bank $\mathbf{M_D}$. 

As for context ($\mathbf{C}$), we model the context-aspect pair in the sentence-pair manner to generate aspect-aware hidden states \cite{AA-LSTM}. The formal input is $\langle$\texttt{[CLS]}; $\mathbf{C}$; \texttt{[SEP]}; $\mathbf{A}$; \texttt{[SEP]}$\rangle$.
%In this case, the hidden states of context words are generated regarding the specific aspect.
%Hence more aspect-related information is contained and aspect-irrelevant information is filtered out in the hidden states.
In this way, we obtain the hidden state of \texttt{[CLS]}: $\mathbf{h_{cls}}$ and a series of aspect-aware context hidden states: $\mathbf{H_C} = \{h_c^i \in \mathbb{R}^{d_e}\}^{N_C}_{i=1}$. %, in which $H_A$ is the hidden state of the aspect's first token.
%As AKEN regards the aspect as a whole concept, the aspect should hold only one position in $H_C$.
%In practice, we take the hidden state of the first token of the aspect as $A_p$.
As BERT has a strong capability of sentence-pair modeling, $\mathbf{h_{cls}}$ contains not only the information from both of the aspect and the context but also their dependencies. Thus we take $\mathbf{h_{cls}}$ as the initial contextualized aspect representation $\mathbf{r_a^0}$.
Then we use $\mathbf{r_a^0}$ to replace the hidden states of aspect words ($\mathbf{H_A} = \{h_a^i \in \mathbb{R}^{d_e}\}^{N_A}_{i=1}$) in $\mathbf{H_C}$, obtaining the initial context memory bank $\mathbf{M_C^0} = {[h_c^1, h_c^2, ..., \mathbf{r_a^0}, ..., h_c^{N}]}$, where $N=N_C -N_A +1$.

$\mathbf{M_D}$ and $\mathbf{M_C^0}$ are two strands of input of KaGR-MN cell.
Along time steps, $\mathbf{M_C}$ is recurrently updated while $\mathbf{M_D}$ remains identical.

\subsection{Knowledge-aware Gated Recurrent Memory Network}
%One core advantage of  KaGRMN-DSG is to leverage the aspect knowledge.
%To this end, we design a KaGR-MN to recurrently integrate the needed aspect knowledge into aspect representation then all context memories.
As the series of context hidden states and description hidden states have been obtained, now the challenge is \textit{how to incorporate as much beneficial aspect knowledge as possible without losing the original semantics obtained from BERT?}.

The first thing is to conserve the original semantics in the context memories obtained from BERT.
To this end, we employ Memory Networks (MNs) as the backbone to store context memories, because that MNs can accurately remember original facts \cite{memory-network}.
Secondly, we are supposed to make the integrated knowledge beneficial.
In other words, we should provide each sample the aspect knowledge it needs.
Hence we propose an aspect-to-description attention (A2D Att) mechanism to summarize the needed aspect knowledge from the description memory bank.
Thirdly, we should integrate the beneficial aspect knowledge into the aspect representation.
Then we propose an adaptive knowledge integration gate, which borrows the idea of gating mechanisms in LSTM \cite{LSTM}.
Gate mechanism has proven its strong ability of information integration in many tasks \cite{AA-LSTM,gru}.
However, only integrate knowledge into aspect representation is insufficient, not exploring the full value of aspect knowledge.
It is intuitive that the aspect knowledge should be incorporated into all context memories.
Besides, an appropriate mechanism should be devised to update the context memory bank.
To achieve these two goals, inspired by Transformer \cite{transformer}, we utilize self multi-head attention to update the context memories, and in the meanwhile, the aspect knowledge in the aspect representation can be spread to all context memories.
Finally, all the above mechanisms form the Knowledge-aware Gated Recurrent Memory Network (KaGR-MN), which combines the advantages of MNs, LSTM, and Transformer.

The architecture of KaGR-MN cell is illustrated in Fig. \ref{fig: mncell}.
In the following texts, we depict the details of KaGR-MN.
%It includes three components: an A2D Att, an AdaKI Gate, and a Self MHA.
%A2D Att dynamically summarizes the needed aspect knowledge.
%AdaKI Gate adaptively integrates the summarized aspect knowledge into $\mathbf{r_a}$.
%Self MHA contextualizes the integrated knowledge and updates $\mathbf{M_C}$.
%In this section, we elaborate on the details of KaGR-MN.
\subsubsection{Dynamic Knowledge Summarizing}
\begin{figure}[t]
 \centering
 \includegraphics[width = 0.48\textwidth]{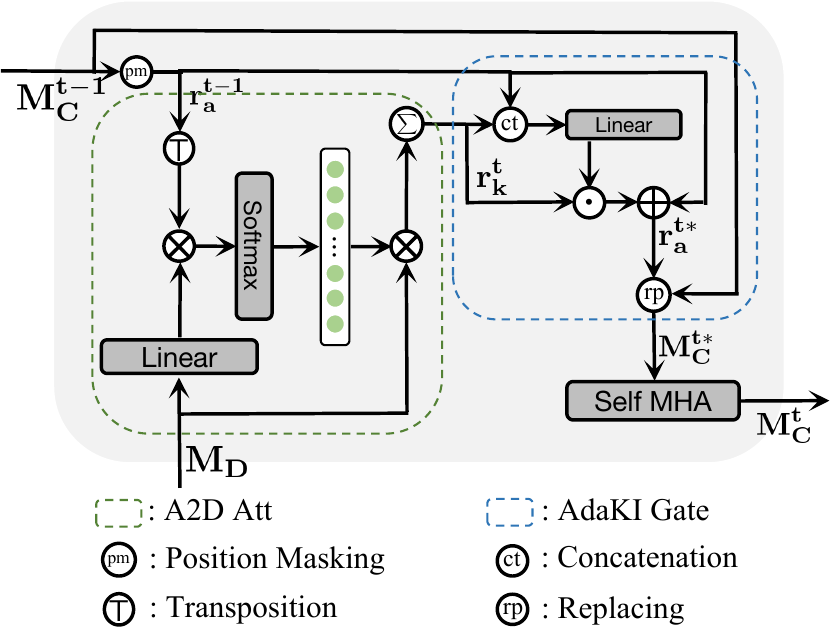}
 \caption{The architecture of KaGR-MN cell.}
 \label{fig: mncell}
\end{figure}

Intuitively, on the one hand, the context-aspect pair of each sample may demand individual aspect knowledge, even if they have the same aspect.
On the other hand, at each time step, KaGR-MN should integrate specifically needed aspect knowledge according to the current cell state.
Therefore, the aspect knowledge summarizing should be dynamic.
To achieve this, we design an aspect-to-description attention (A2D Att) mechanism to dynamically summarize the specifically needed aspect knowledge from the description memory bank $\mathbf{M_D}$ at each time step.
The architecture of A2D Att is shown in Fig. \ref{fig: mncell}.
At each time step ($t$), the aspect representation of previous time step $\mathbf{r_a^{t-1}}$ serves as the cell state and is used to query $\mathbf{M_D}$.
Then an attention weight $\alpha$ is assigned to each $h_d$ regarding its importance to $\mathbf{r_a^{t-1}}$:
% $\alpha_i$  is produced as:
\begin{align}
\begin{split}
\alpha_i=&\texttt{SoftMax} (\mathcal{F}(h_d^i, r_a^{t-1})) \\
        =&\frac {\exp(\mathcal{F}(h_d^i, r_a^{t-1}))}{\sum_{k=1}^{N_D} \exp(\mathcal{F}(h_d^k, r_a^{t-1}))} 
\end{split}
\end{align}
where $\mathcal{F}(h_d^i, r_a^{t-1}))$ is a score function defined as:
\begin{equation}
\mathcal{F}(h_d^i, r_a^{t-1}))=(\mathbf{{W_d}}\,{h_d^i} +\mathbf{b_d})\,(\mathbf{{r_a^{t-1}}})^\mathsf{T} \label{eq: f}
\end{equation}
where $\mathbf{W_d}$ and $\mathbf{b_d}$ are weight matrix and bias respectively, and $\mathsf{T}$ denotes transposition.
Then we can obtain the summarized knowledge representation as: $\mathbf{r_k^t}=\sum_{i=1}^{N_D}\alpha_i h_d^i$.
\subsubsection{Adaptive Knowledge Integration}
As the specifically needed knowledge has been summarized, it should be integrated into the aspect representation regarding the current cell state.
As the gate mechanisms \cite{LSTM,gru,lightgate} have proven their ability of controlling information flow,
here we design an \textbf{Ada}ptive \textbf{K}nowledge \textbf{I}ntegration (AdaKI) Gate to integrate $\mathbf{r_k^t}$ into $\mathbf{r_{a}^{t-1}}$.
Its architecture is shown in Fig. \ref{fig: mncell}.
AdaKI Gate can be formulated as:
\begin{equation}
\mathbf{r_{a}^{t*}} = \mathbf{r_{a}^{t-1}}+ \mathbf{r_k^t} \odot (\mathbf{W_{k}} [\mathbf{r_{a}^{t-1}}, \mathbf{r_k^t}])
\end{equation}
where $\odot$ denotes Hadamard product, $[,]$ denotes concatenation and $\mathbf{W_k}$ is weight matrix.
The core of AdaKI Gate is to produce a gate vector using $\mathbf{r_k^t}$ and $\mathbf{r_{a}^{t-1}}$.
This gate vector achieves the fine-grained control on each dimension of $\mathbf{r_k^t}$.

There are two merits of this fine-grained control.
First, AdaKI Gate can determine what knowledge and how much knowledge from $\mathbf{r_k^t}$ should be integrated into $\mathbf{r_{a}^{t-1}}$.
Second, it can map the integrated knowledge into the same semantic space of $\mathbf{r_{a}^{t-1}}$ and $\mathbf{M_{C}^{t-1}}$.
This adaption helps maintain the semantics consistency of $\mathbf{r_{a}^{t*}}$ and $\mathbf{M_{C}^{t-1}}$, which is beneficial to later knowledge contextualizing.
%On the other hand, the semantic space adaption preserves the original contextual information contained in $r_{A}^{t-1}$, e.g. dependencies between the aspect and related context words.
In Sec. \ref{sec: gate}, we investigate the effect of different knowledge gates used here.
%To convert $\widetilde{r_A}$ into the common space of $H_C$, we project it as: $r_A = W_p \widetilde{r_A}$. % in which $W_p$ is trainable weight.
After $\mathbf{r_{a}^{t*}}$ is obtained, it replaces $\mathbf{r_{a}^{t-1}}$ in $\mathbf{M_{C}^{t-1}}$, forming $\mathbf{M_{C}^{t*}}$.
%Our method regards the aspect as a whole concept and it only takes one place in $H_C$.

\subsubsection{Knowledge Contextualizing and Context Memory Bank Updating}
Although the needed beneficial knowledge has been integrated into $\mathbf{r_{a}^{t*}}$, the other context memories in $\mathbf{M_{C}^{t*}}$ remain the same as the ones in $\mathbf{M_{C}^{t-1}}$.
Intuitively, all context memories should benefit from aspect knowledge to facilitate aspect-related information aggregation.
%Intuitively, the knowledge should be integrated into not only the aspect representation but also each context memory.
To achieve this, we propose a knowledge contextualizing mechanism to broadcast the newly-integrated knowledge in $\mathbf{r_{a}^{t*}}$ to all context memories in $\mathbf{M_{C}^{t*}}$.
Here we borrow the idea of self-attention \cite{self-iclr,transformer}, which can effectively relate the different tokens in a sentence and capture the intra-sentence dependencies.

In this work, we adopt the self multi-head attention (Self MHA) formulation in \cite{transformer}.
We first map $\mathbf{M_{C}^{t*}}$ to queries ($\mathbf{Q}$), keys ($\mathbf{K}$) and values ($\mathbf{V}$) matrices by individual linear projections, where $\mathbf{Q}, \mathbf{K}, \mathbf{V} \in \mathbb{R}^{N\times d_s}$.
And this process repeats $H^s_n$ times, where $H^s_n$ is the number of heads and $H^s_n \times d_s = d_e$.
The scaled dot-product attention is used to produce the output of each head, then all of the $H^s_n$ outputs are concatenated to form the updated context memory bank $\mathbf{M_C^t}$:
\begin{equation}
\mathbf{M_C^t} = \|_{h=1}^{H^s_n} \texttt{SoftMax}\left(\frac{\mathbf{Q} \mathbf{K}^{\mathsf{T}}}{\sqrt{d_{s}}}\right) \mathbf{V}
\end{equation}
This simple but effective knowledge contextualizing mechanism updates $\mathbf{M_{C}^{t*}}$ and $\mathbf{r_{a}^{t*}}$ by letting context memories (including $\mathbf{r_{a}^{t*}}$)  exchange useful information with each other, which is beneficial to capture aspect-related information.
Along time steps, $\mathbf{r_{a}}$ and $\mathbf{M_{C}}$ would contain more and more reasonable and beneficial semantics for ASC.

\subsection{Dual Syntax Graph Network}
As proven in prior works, GCN can capture local syntactic information and Relational MHA can capture the global relation between the aspect and each context word via operating on the star-shaped aspect-oriented syntax graph (as shown in Fig. \ref{fig: framework}).
However, as we have discussed in Sec. \ref{sec:introduction}, they have respective shortages.
They can only capture one of the two kinds of syntactic information and lose the other.
Previous models only employ one of them, leading to insufficient syntactical information.

%This paper proposes to marry them together to compensate for each other's shortage.
%However, in previous works, only one of them was adopted, lack of their combining and interaction.
To this end, we propose DSG-Net (as shown in Fig. \ref{fig: framework}) which marries the proposed Position-aware GCN and Relational MHA and learns their interaction, capturing sufficient syntactic information.
%captures sufficient syntactical information by combining GCN and Relational MHA then lear.
%Hence, the proper aspect knowledge can flow in more sufficient dependency paths, facilitating the aspect sentiment reasoning.

\subsubsection{Local Syntactic Information Modeling}
%Our framework regards the aspect as a whole concept and $r_A$ only has one token place in $H_C$, while the vanilla syntax graph $G$ contains all of the aspect words nodes and their edges.
\textbf{Graph Construction}
Based on the original syntax graph $G$\footnote{obtained by spaCy toolkit: https://spacy.io/}, %, following previous works\cite{asgcn,DGEDT}},
%before feeding $\mathbf{M_C^T}$ into GCN.
% because the vanilla syntax graph $G$ contains all of the aspect words while $r_A$ only holds one position in $H_C$.  
we first add a new aspect node $A$ and merge all edges between nodes of aspect words and non-aspect context words to $A$.
Then we delete all of the original nodes of aspect words and their edges. The obtained graph is similar to $G$, and only several context word nodes are connected to $A$.
Thus we term it sparse graph $G_s$ (shown in Fig.\ref{fig: framework}).

\noindent \textbf{Position-aware GCN}
In this work, we augment the standard GCN with a position weight $w_p^i = 1 - \frac {\vert i-\tau\vert}{N+1}$, in which $\tau$ denotes the position of aspect, $i$ denotes the $i^{th}$ context word.
As the Self MHA in KaGR-MN does not consider the order of context memories, some positional and ordering information may be lost. $w_p^i$ can supplement this information, which helps capture local syntactic information.
%There are three reasons why we use $w_p^i$.
Besides, it indicates the position of $A$ and highlights the potential aspect-related words which are generally closer to $A$.
%relational MHA does not need these information because each non-aspect context words are non-adjenct.\\
%To indicate the position of $A$ and highlight the potential aspect sentiment trigger words, 
In $l^{th}$-layer, the local neighborhood information is aggregated as: %$l^{th}$-layer node update function is:
\begin{equation}
h_i^l= \sum_{j\in \mathcal{N}_{i}^s} \mathbf{W_g^s} (w_p^j\ h_j^{l-1})  /(d_i + 1) + \mathbf{b_g^s}
\end{equation}
in which $\mathcal{N}_{i}^s$ is the first-order neighbors of node $i$ (including $i$) in $G_s$, $d_i$ is the degree of node $i$, $\mathbf{W_g^s}$ and $\mathbf{b_g^s}$ are  weight matrix and bias.

\subsubsection{Global Relational Information Modeling}
%Relational MHA is the core of the recent proposed R-GAT+BERT \cite{RGAT} model.
We obtain the star-shaped aspect-oriented syntax graph
%first reshape and prune the syntax graph\footnote{generated by the Biaffine Parser \cite{biaffine}, following \cite{RGAT}.} 
following% the strategy in
\cite{RGAT}.
In this syntax graph, every context word directly connects to the aspect node $A$, so we term it dense graph $G_d$.
Then we employ the Relational MHA to model the global relational dependency between aspect and each context word.
The node representation is:
\begin{equation}
\begin{aligned}
h_{i}&=\sum_{m=1}^{H^d_n} \Bigg(\sum_{j \in \mathcal{N}_{i}^d} \beta_{ij}^m \mathbf{W_{m}^{1}} h_{j}\Bigg)/H^d_n \\
%\beta_{i j}^k&=\frac{\texttt{exp} (g_{i j}^n)}{\sum_{j=1}^{\mathcal{N}_{i}^d} \texttt{exp} (g_{i j}^k)}\\
\beta_{i j}^m& = \texttt{SoftMax}(g_{ij}^m)\\
g_{i j}^m&=\texttt{ReLU} \left(r_{i j} \mathbf{W_{m}^2}+\mathbf{b_{m}^1}\right) \mathbf{W_{m}^3}+\mathbf{b_{m}^2} 
\end{aligned}
\end{equation}
where $H_n^d$ denotes the head number, $r_{ij}$ is the embedding of the relation between nodes $i$ and $j$, $\mathbf{W_{m}^{1,2,3}}$ and $\mathbf{b_{m}^{1,2}}$ are weight matrices and biases. % where $* \in \{\mathbf{r1}, \mathbf{r2}, \mathbf{r3}\}$.

\subsubsection{Dual Syntactic Information Fusion}
Now the Position-aware GCN has captured the important local syntactic information and the Relational MHA has captured the important global relational information.
To integrate them together and let them compensate for each other, we concatenate the aspect node representations respectively derived by Position-aware GCN and Relational MHA, then we employ a multi-layer perception (MLP), which can automatically abstract the integrated representation \cite{cvprmlp,dcrnet}, to generate the unified node representation sequence, which include the unified aspect representation $\mathbf{\widetilde{R_a}}$.

\subsection{Knowledge Re-enhancement\\}
After graph modeling, sufficient syntactic information has been integrated into $\mathbf{\widetilde{R_a}}$.
On the one hand, some new clues may be captured by DSG-Net and retained in $\mathbf{\widetilde{R_a}}$.
Thus $\mathbf{\widetilde{R_a}}$ may further need more aspect knowledge to collaborate with these new clues to support ASC.
On the other hand, as the syntax graph may be imperfectly generated by the parser, some wrong connections and relations may be introduced.
In this case, re-integrating some knowledge can help alleviate the influence of the imperfect syntax graph.
To this end, we design a knowledge integrating gate (KI Gate) to re-enhance $\mathbf{\widetilde{R_a}}$ with further needed knowledge contained in $\mathbf{r_k^T}$.
The function of KI gate is given as:
\begin{equation}
\mathbf{R_a} = \mathbf{\widetilde{R_a}} + \mathbf{r_k^T} * \mathbf{W_{k}^r}[\mathbf{\widetilde{R_a}},\mathbf{r_k^T}]
\end{equation}
where $\mathbf{W_{k}^r}$ is weight matrix.
%Note that KI Gate is different from AdaK Gate.
Here $\mathbf{\widetilde{R_a}}$ and $\mathbf{r_k^T}$ produces a gate scalar rather than a gate vector.
There is no subsequent contextualizing module thus $\mathbf{r_k^T}$ can be directly integrated into $\mathbf{\widetilde{R_a}}$ without fine-tuning for adaption.
In Sec. \ref{sec: gate}, we investigate the effect of different knowledge gates used here.

%where $\odot$ is an element-wise product operator.
%R_A^K$ can provide rich aspect knowledge such as the properties and meaning of the aspect.
%AKRgate has two main merits.
%The first merit is that AKRgate can flexibly reach a trade-off between the information from $\widetilde {R_A}$ and the aspect knowledge from $R_K^A$.
%The other merit is that AKRgate shortens the path that the loss backpropogates to $Description\ BERT$.
\subsection{Aspect-related Semantics Aggregation}
Here we employ an Aspect-to-Context Attention (A2C Att) mechanism to aggregate the aspect-related semantics retained in all hidden states into a final representation $\mathbf{R_f}$.
Similar to A2D Att, A2C Att can be formulated as:
\begin{align}
\beta_i=&\texttt{SoftMax} (\mathcal{F}(h_c^i, \mathbf{R_a})) \\
\mathcal{F}(h_c^i, R_a)&=(\mathbf{W_{ac}}\,{h_c^i} +\mathbf{b_{ac}})\,(\mathbf{R_a})^\mathsf{T} \\
\mathbf{R_f}=&\sum_{i=1}^{N}\alpha_i h_c^i
\end{align}
where $\mathbf{W_{ac}}$ and $\mathbf{b_{ac}}$ are weight and bias.

\subsection{Sentiment Classification\\}
%Having obtained the final aspect representation $R_a$, 
We concatenate $\mathbf{R_f}$ with $\mathbf{h_{cls}}$ and then fed the final vector into a linear layer, which is followed by a $\texttt{SoftMax}$ classifier for prediction:
\begin{equation}
\mathbf{P} = \texttt{SoftMax}(\mathbf{W_p}[\mathbf{h_{cls}},\mathbf{R_f}] + \mathbf{b_p})
\end{equation}
where $\mathbf{P}$ is the predicted sentiment distribution, $\mathbf{W_p}$ and $\mathbf{b_p}$ are weight matrix and bias.
The cross-entropy loss function is adopted for model training.

There are two reasons why we introduce $\mathbf{h_{cls}}$ here.
First, this can add a skip connection to BERT, shortening its loss back-propagation path to facilitate training.
The second is for robustness.
Possibly the syntax graphs are imperfect and the integrated knowledge contains noise.
Hence $\mathbf{h_{cls}}$ serves as a reference and makes the whole model more robust.
%\paragraph{Loss Function\\}

\begin{comment}
\begin{equation}
%\setlength{\abovedisplayskip}{1pt}
%\setlength{\belowdisplayskip}{3pt}
\mathcal L=-\sum_{(s,y)\in\mathcal S} log(p_y)% + \lambda\|\Theta\|_{2}
\end{equation}
where $S$ denotes all training samples, $s$ is the input sample and $y$ is the ground-truth label.
\end{comment}

%% file: experiment.tex
% !TEX root = bare_jrnl.tex
\section{Experiments}
\subsection{Datasets}
We conduct experiments on three popular datasets for the ASC task:
%The datasets are of laptop (Lap) and restaurant (Res) domains.
Lap14 and Res14 datasets are from SemEval 2014 task 4 \cite{semeval2014}, and Res15 dataset is from SemEval 2015 task 12 \cite{semeval2015}.
The statistics of all datasets are presented in Table \ref{table: dataset}.
\begin{table}[ht]
\centering
\caption{Dataset statistics of the three datasets.}
\fontsize{8}{10}\selectfont
\setlength{\tabcolsep}{2.5mm}{
\begin{tabular}{ccccccc}
\toprule
\multirow{2}{*}{Dataset} & \multicolumn{2}{c}{Positive} & \multicolumn{2}{c}{Neutral} & \multicolumn{2}{c}{Negative} \\\specialrule{0em}{0pt}{1pt} \cline{2-3} \cline{4-5} \cline{6-7}\specialrule{0em}{1pt}{1.5pt}
                         & Train         & Test         & Train         & Test        & Train         & Test         \\ \specialrule{0em}{0pt}{1pt}\hline \specialrule{0em}{1.5pt}{1.5pt}
Lap14                    & 994           & 341          &  464          &    169      &  870          &    128       \\
Res14                    & 2164          & 728          & 637           & 196         & 807           & 196          \\
Res15                    & 912           & 326          & 36            & 34          & 256           & 182         \\\bottomrule
\end{tabular}}
\label{table: dataset}
\end{table}
% \usepackage{multirow}

\subsection{Experiment Setup}
%Following previous works, the Biaffine Parser \cite{biaffine} is used for generating the dependency graph of Relational MHA \cite{RGAT}, and spaCy toolkit is used to generate the dependency graph
We adopt the BERT-base uncased version \cite{bert}. % and the PyTorch BERT implementation \cite{transformers} is used.
We train our model using Adam optimizer \cite{Adam} with default configuration.
%The learning rate set is $[5e^{-5}, 2e{-5}, 1e{-5}]$ and the weight decay set is $[0.5, 0.05, 0.002]$.
The hyper-parameters are listed in Table \ref{table: hyper-param}.
%About 70\% aspects are equipped with non-null descriptions retrieved from DBpedia/Wikipedia.
%The original syntax graphs of GCN and Relational MHA are obtained by spaCy toolkit and Biaffine Parser\cite{biaffine}, following \cite{asgcn} and \cite{RGAT}.
\begin{table}[t]
\fontsize{9}{11}\selectfont
\centering
\caption{Setting of hyper-parameters.}
\begin{tabular}{c|ccc}
\toprule
\multirow{2}{*}{Hyper-params} & \multicolumn{3}{c}{Dataset} \\\cline{2-4}
                              & Lap14   & Res14   & Res15   \\ \midrule
         learning rate        & $1\times 10^{-5}$  &   $5\times 10^{-5}$      & $3 \times 10^{-5}$        \\
          batch size          & 32     &   32    &   32    \\
            dropout rate     &  0.3    &    0.3  &   0.3   \\
           $d_e$              &  768    &  768    &  768   \\ 
$d_s$                 &  256     &    256     &   128      \\
$H_n^s$                 &  3     &   3     &     6    \\
$H_n^d$                &  2     &   4    &    6     \\
$\mathbf{T}$                 &   4    &      4   &    2    \\
GCN layer number             &   2      &    2     &   2      \\
           \bottomrule
\end{tabular}
\label{table: hyper-param}
\end{table}
Accuracy (Acc) and Macro-F1 (F1) are adopted as evaluation metrics.
As there is no official validation set, following previous works \cite{asgcn,bigcn,DGEDT}, we run our model three times with random initialization and report the average results on test sets, as shown in Table \ref{table: avg results}.
And to compare with the works reporting best results, we also report the best results on test sets, as shown in Table \ref{table: best results}.
All computations are done on an NVIDIA Quadro RTX 6000 GPU.

\begin{comment}
Besides standard KaGRMN-DSG (M$_0$), we also evaluate some variants (M$_{1}$\textasciitilde M$_4$) of M$_0$ to empirically verify its effectiveness from different perspectives.
M$_1$: DSG-Net is evaluated as a single model.
M$_2$: we replace the input description with the aspect so no aspect knowledge is leveraged.
M$_3$: KaGR-MN is evaluated as a single model.
M$_4$: KI Gate is removed.
%In M$_1$ and M$_3$, the BERT encoder and sentiment classification module remain the same.
\end{comment}
%We compare our models with some state-of-the-art (SOTA) baselines, as listed in Table \ref{table: avg results}.
% 
\subsection{Compared Baselines}
According to what kinds of external information are utilized, we divide the baselines into several group:\\
1) No external information is used:
 \begin{itemize}
 \item IAN \cite{IAN} separately encodes the aspect and context, then model their interactions using an interactive attention mechanism.
\end{itemize}
2) External corpus is used:
 \begin{itemize}
 \item  PRET+MULT \cite{hrd} first pre-trains the model on document-level task, then trains the model on both document-level sentiment classification and ASC in the multi-task learning framework.
 \item TransCap \cite{transcap} utilizes a devised aspect-based capsule network to transfer knowledge from document-level task to aspect-level task.
\end{itemize}
3) Syntax Graph is used: 
 \begin{itemize}
 \item ASGCN \cite{asgcn} employs a GCN to encode the syntax graph for capturing local syntactic information.
 \item BiGCN \cite{bigcn} convolutes over hierarchical syntactic and lexical graphs to encode not only original syntactic information but also the corpus level word co-occurrence information.
\end{itemize}
4) BERT encoder is used: 
 \begin{itemize}
 \item BERT-SPC \cite{bert} takes the same input as our model and use $h_{cls}$ for sentiment classification.
 \item AEN-BERT \cite{aen-bert} adopts BERT encoder and uses the attentional encoder network to model the interactions between the aspect and context.
\end{itemize}
5) Both of syntax graph and BERT encoder are used: 
  \begin{itemize}
 \item R-GAT+BERT \cite{RGAT} use the relational graph attention network to aggregate the global relational information from all context word into the aspect node representation.
 \item  DGEDT-BERT \cite{DGEDT} employs a dual-transformer network to model the interactions between the flat textual knowledge and dependency graph empowered knowledge. 
 \item A-KVMN+BERT \cite{kvmn-eacl} uses a key-value memory network to leverage not only word-word relations but also their dependency types.
 \item BERT+T-GCN \cite{tgcn} leverages the dependency types in T-GCN and use an attentive layer ensemble to learn the comprehensive representation from different T-GCN layers.
 \item SAGAT \cite{sagat} utilizes graph attention network and BERT to fully obtain both syntax and semantic information.
 \item KGCapsAN-BERT \cite{kgcap} utilizes multi-prior knowledge to guide the capsule attention process and use a GCN-based syntactic layer to integrate the syntactic knowledge.
\end{itemize}

And we label all models with what kinds of external information they leverage, as shown in Table \ref{table: avg results} and Table \ref{table: best results}.

\subsection{Main Results}

\begin{table*}[ht]
\fontsize{8}{10}\selectfont
\centering
\caption{Performances comparisons of average results with random initialization. $\mathcal{K}, \mathcal{B}, \mathcal{T}$ and $\mathcal{G}$ denote the model leverages aspect $\mathcal{K}$nowledge, $\mathcal{B}$ERT, extra $\mathcal{T}$raining corpus and syntax $\mathcal{G}$raph, respectively. Best results are in \textbf{bold} and previous SOTA results are \underline{underlined}. 
%$^\natural$ denotes the results are retrieved from \cite{asgcn}.
$^*$ denotes that we produce the results using their original source codes.
$^\dag$ indicates KaGRMN-DSG significantly outperforms baselines under t-test ($p<0.01$).}
\setlength{\tabcolsep}{3mm}{
\begin{tabular}{ccccccccccccc}\toprule
\multirow{2}{*}{\tabincell{c}{External \\ Information}}  & \multirow{2}{*}{Model} & \multicolumn{2}{c}{Lap14} & \multicolumn{2}{c}{Res14} & \multicolumn{2}{c}{Res15}  \\\cline{3-8} 
                                         &                               &Acc       &F1        &Acc        &F1        &Acc        &F1       \\\midrule
%$\mathcal{-} \quad \mathcal{-} \quad \mathcal{-}$  & LSTM$^\natural$ \cite{TDLSTM}   &69.28     &63.09     &78.13     &67.47  &77.37       &55.17        \\
$\mathcal{-} \quad \mathcal{-} \quad \mathcal{-}$  & IAN \cite{IAN}       &72.05     & 67.38    &79.26    &70.09  & 78.54    & 52.65         \\
$\mathcal{-} \quad \mathcal{T} \quad \mathcal{-}$  & PRET+MULT \cite{hrd} &71.15     & 67.46    &79.11   & 69.73  & 81.30    & 68.74       \\
$\mathcal{-} \quad \mathcal{T} \quad \mathcal{-}$& TransCap \cite{transcap} & 73.51  & 69.81    & 79.55    & 71.41  & -      & -      \\ 
$\mathcal{-} \quad \mathcal{-} \quad \mathcal{G}$& ASGCN \cite{asgcn}     & 75.55    &71.05     &80.77    & 72.02  & 79.89   & 61.89        \\
%$\mathcal{-} \quad \mathcal{-} \quad \mathcal{G}$& CDT\cite{CDT}      & 77.19       & 72.99    & 82.30    & 74.02   & -      & -           \\
%$\mathcal{-} \quad \mathcal{-} \quad \mathcal{-}$& TNet-ATT(+AS)  & 77.62       & 73.84       & 81.53        & 72.90       & -            & -           \\
$\mathcal{-} \quad \mathcal{-} \quad \mathcal{G}$& BiGCN \cite{bigcn}    & 74.59    &71.84    &81.97     &73.48    & 81.16    & 64.79       \\
%$\mathcal{-} \quad \mathcal{-} \quad \mathcal{G}$& RepWalk \cite{walk}    & 78.2      & 74.3        & 83.8      & 76.9      & -          & -    \\
$\mathcal{-} \quad \mathcal{B} \quad \mathcal{-}$& BERT-SPC$^*$ \cite{bert} & 78.47    &73.67  &84.94   & 78.00   & 83.40    &  65.00   \\
$\mathcal{-} \quad \mathcal{B} \quad \mathcal{-}$& AEN-BERT \cite{aen-bert} & 79.93    &76.31  &83.12   & 73.76   & -    &  -   \\
$\mathcal{-} \quad \mathcal{B} \quad \mathcal{G}$& R-GAT+BERT$^*$ \cite{RGAT} & 79.31    &75.40  &86.10   & \underline{80.04}   & 83.95    &  69.47   \\
$\mathcal{-} \quad \mathcal{B} \quad \mathcal{G}$& DGEDT-BERT \cite{DGEDT}   & 79.8    &75.6    &\underline{86.3}      & 80.0   & 84.0   &71.0    \\
%$\mathcal{-} \quad \mathcal{B} \quad \mathcal{G}$& A-KVMN+BERT \cite{kvmn-eacl}   & 79.78    &76.14    &85.98      & 77.94   & \underline{84.14}   &68.49    \\

 $\mathcal{-} \quad \mathcal{B} \quad \mathcal{G}$ & A-KVMN+BERT$^*$ \cite{kvmn-eacl}   & 79.20    &75.76   &85.89      & 78.29   & 83.89  &67.88    \\
 $\mathcal{-} \quad \mathcal{B} \quad \mathcal{G}$ & BERT+T-GCN$^*$ \cite{tgcn}    & \underline{80.56}    &\underline{76.95}    &85.95      & 79.40   & \underline{84.81} & \underline{71.09} \\\hline
$\mathcal{K} \quad \mathcal{B} \quad \mathcal{G}$ & KaGRMN-DSG (Ours)    &\textbf{81.87$^\dag$}    & \textbf{78.43$^\dag$}     &\textbf{87.35$^\dag$}   & \textbf{81.21$^\dag$}      &\textbf{86.59$^\dag$}   &\textbf{74.46$^\dag$}        \\
   & Our Improvements      & \textbf{1.62\%}  &\textbf{1.92\%} & \textbf{1.22\%} &\textbf{1.46\%} &\textbf{2.10\%} &\textbf{4.74\%} \\
\bottomrule
\end{tabular}}
\label{table: avg results}
\end{table*}

\begin{table*}[ht]
\fontsize{8}{10}\selectfont
\centering
\caption{Performances comparisons of best results. $\mathcal{K}, \mathcal{B}, \mathcal{T}$ and $\mathcal{G}$ denote the model leverages aspect $\mathcal{K}$nowledge, $\mathcal{B}$ERT, extra $\mathcal{T}$raining corpus and syntax $\mathcal{G}$raph, respectively. Best results are in \textbf{bold} and previous SOTA results are \underline{underlined}. $^*$ denotes that we produce the results using their original source codes.}
%}
\setlength{\tabcolsep}{3mm}{
\begin{tabular}{ccccccccccccc}\toprule
\multirow{2}{*}{\tabincell{c}{External \\ Information}}  & \multirow{2}{*}{Model} & \multicolumn{2}{c}{Lap14} & \multicolumn{2}{c}{Res14} & \multicolumn{2}{c}{Res15}  \\\cline{3-8} 
                                         &                               &Acc       &F1        &Acc        &F1        &Acc        &F1       \\\midrule
%$\mathcal{-} \quad \mathcal{-} \quad \mathcal{-}$  & LSTM$^\natural$ \cite{TDLSTM}   &69.28     &63.09     &78.13     &67.47  &77.37       &55.17        \\
$\mathcal{-} \quad \mathcal{B} \quad \mathcal{-}$& BERT-SPC$^*$ \cite{bert} & 78.84    &73.95 &85.80   & 78.48   & 83.76    &  68.33   \\
$\mathcal{-} \quad \mathcal{B} \quad \mathcal{G}$& SAGAT \cite{sagat} & 80.37    &76.94  &85.08   & 77.94   & -    &  -   \\
$\mathcal{-} \quad \mathcal{B} \quad \mathcal{G}$& KGCapsAN-BERT \cite{kgcap} & 79.47    &76.61  &85.36   & 79.00   & -    &  -   \\
$\mathcal{-} \quad \mathcal{B} \quad \mathcal{G}$& R-GAT+BERT$^*$ \cite{RGAT} & 79.46   &75.75  &\underline{86.61}   & \underline{80.78}   & 84.13   &  71.12   \\
$\mathcal{-} \quad \mathcal{B} \quad \mathcal{G}$& A-KVMN+BERT \cite{kvmn-eacl}   & 79.78    &76.14    &85.98      & 77.94   & 84.14   &68.49    \\
 $\mathcal{-} \quad \mathcal{B} \quad \mathcal{G}$ & BERT+T-GCN \cite{tgcn}    & \underline{80.88}    &\underline{77.03}    &86.16      & 79.95   & \underline{85.26} & \underline{71.69} \\\hline
$\mathcal{K} \quad \mathcal{B} \quad \mathcal{G}$ & KaGRMN-DSG (Ours)    &\textbf{82.13}    & \textbf{79.42}     &\textbf{87.68}  & \textbf{81.98}      &\textbf{87.08}   &\textbf{75.34}        \\
   & Our Improvements      & \textbf{1.55\%}  &\textbf{3.10\%} & \textbf{1.24\%} &\textbf{1.49\%} &\textbf{2.13\%} &\textbf{5.09\%} \\
\bottomrule
\end{tabular}}
\label{table: best results}
\end{table*}

The performance comparison of all models on average scores is shown in Table \ref{table: avg results}, and the comparison on best scores is shown in Table \ref{table: best results}.
We can observe that:
Syntax graphs, external training corpus, and BERT can all improve ASC.
Especially, simple BERT-SPC significantly outperforms all models that do not adopt BERT, even if some of them leverage syntax graph and external training corpus.
This shows the power of pre-trained language models on ASC.
And combining BERT and syntactic information can further improve results as sufficient semantics captured by BERT and the syntactic information conveyed by syntax graphs can cooperate to assist ASC.
However, all baselines do not leverage aspect knowledge and only consider either local syntactic information or global relational information.
As a result, their derived aspect representation lack some important clues of aspect and their captured syntactic information is insufficient, leading to their inferior performance compared to our KaGRMN-DSG model.
%This proves that the contribution of the aspect knowledge is nontrivial and distinct.

We obtain consistent improvements over baselines in terms of Acc and F1 on all datasets, achieving new state-of-the-art results.
%Compared with DGEDT-BERT which is the best-performing baseline, 
On average results, our KaGRMN-DSG overpasses previous best results by 1.92\%, 1.46\%, and 4.74\% in terms of Macro-F1 on Lap14, Res14, and Res15 datasets respectively.
On best results, KaGRMN-DSG overpasses previous best results by 3.10\%, 1.49\%, and 5.09\% in terms of Macro-F1 on Lap14, Res14, and Res15 datasets respectively.
The improvements are contributed by the superiorities of KaGR-MN, which effectively leverage beneficial aspect knowledge, and DSG-Net, which combines GCN and Relational MHA to capture sufficient syntactic information.
%On Lap14 dataset, compared with SAGAT which reports the best scores, our average scores overpass their best ones by 1.1\% on Acc and 1.2\% on F1.
%Compared with DGEDT-BERT, on Res14 dataset, we achieve improvements of 0.8\% on Acc and 1.2\% on F1.
%Remarkably, on Res15 dataset, we achieve improvements of 2.4\% on Acc and 3.1\% on F1.

\subsection{Ablation Study}
\begin{table*}[t]
\fontsize{8}{11}\selectfont
\centering
\caption{Results of ablation study.}
\setlength{\tabcolsep}{3mm}{
\begin{tabular}{l|ccc}\toprule
\multirow{2}{*}{Variants} & Lap14 & Res14 & Res15  \\\cline{2-4}
               &Acc               &Acc                &Acc              \\\midrule
    M$_0$: KaGRMN-DSG (full model) &\textbf{81.87}     &\textbf{87.35 }     &\textbf{86.59} \\ \hline
     M$_1$: w/o Aspect Knowledge (descriptions are replaced with aspects)    &80.30 ($\downarrow1.57$)  &86.43  ($\downarrow0.92$)  &85.24  ($\downarrow1.35$)   \\
M$_2$: only KaGRMN (w/o DSG-Net + KI Gate + A2C Att)   &80.72 ($\downarrow1.15$)  &86.55  ($\downarrow0.80$)    &85.36  ($\downarrow1.23$)    \\ \hline
  M$_3$:  w/o DSG-Net   &80.56 ($\downarrow1.31$)  &86.43  ($\downarrow0.92$)  &85.56 ($\downarrow1.03$)\\
    M$_4$: w/o Relational MHA     & 80.98 ($\downarrow0.89$)   &86.67 ($\downarrow0.68$)    &85.79  ($\downarrow0.8$)    \\
   M$_5$: w/o Position-aware GCN    & 81.03 ($\downarrow0.84$)   &86.76 ($\downarrow0.59$)   &85.67   ($\downarrow0.92$) \\ \hline
 M$_6$: w/o KI Gate    &81.09 ($\downarrow0.78$)   &87.11 ($\downarrow0.24$)    &85.79    ($\downarrow0.80$)    \\ \hline
M$_7$: w/o A2C Att   &81.50 ($\downarrow0.37$)     &87.00 ($\downarrow0.35$)   &86.41  ($\downarrow0.18$)     \\ \hline
    M$_8$: w/o  A2D Att &80.93 ($\downarrow0.94$)  & 86.70 ($\downarrow0.65$)  &85.61    ($\downarrow0.98$)      \\
    M$_9$: w/o Self MHA   &  80.36 ($\downarrow1.51$)  & 86.46 ($\downarrow0.89$)   &85.24  ($\downarrow1.35$)    \\

\bottomrule
\end{tabular}}
\label{table: ablation}
\end{table*}
We empirically analyze KaGRMN-DSG and prove the necessity of every component by conducting an ablation study, whose results are shown in Table \ref{table: ablation}.
In this section we answer the following research questions (RQs):

\textit{Effect of Aspect Knowledge.}
To study the pure impact of aspect knowledge, we devise two variants: M$_1$ and M$_2$. 
In M$_1$, the original description is replaced with the aspect itself.
In this case, there is no aspect knowledge available for KaGR-MN and its function becomes modeling the interactions between the aspect and context.
Surprisingly, even without knowledge, M$_2$ can obtain promising results. 
We attribute this to the advanced architecture and effective functions of KaGR-MN, in which aspect and context are separately encoded and their interactions are effectively modeled by KaGR-MN.
On the other hand, the performance degradation of M$_0$ convincingly demonstrates the pure improvements contributed by the aspect knowledge conveyed by aspect descriptions.
In M$_2$, DSG-Net, KI Gate, and A2C Att are all removed, so M$_2$ has a BERT+KaGR-MN architecture and the final aspect representation is used for prediction.
M$_2$ consistently outperforms baselines, proving that KaGR-MN can derive a good enough aspect representation in which the clues for aspect sentiment reasoning are retained.
Along time steps, recurrently leveraging aspect knowledge, KaGR-MN can capture more and more beneficial clues, semantics and dependencies then retain them in aspect representation and context memories.
And effectively utilizing beneficial aspect knowledge is the key advantage of our method compared with previous works.

\textit{Effect of Syntactic Information.}
%The competitive results of M$_1$ show DSG-Net is effective as a single model.
The results gap of M$_3$ and M$_0$ shows the improvement DSG-Net achieves by cooperating with the aspect knowledge.
These results validate the advantages of combining both kinds of syntactic information to capture sufficient syntactic information.
We then study the effects of Position-aware GCN and Relational MHA.
We can observe that both M$_4$ and M$_5$ perform worse than M$_0$, proving both the local syntactic information and global relational information should be captured for ASC.
In previous works, only either one of them is considered, leading to insufficient syntactic information.
In contrast, our model marries them and lets them compensate for each other, sufficiently capturing syntactic clues.
%And their mutual interaction is learned to capture sufficient syntactic information.
%This demonstrates that KAGR-MN can effectively leverage the aspect knowledge for reasoning aspect sentiment.

%M$_2$ shares the same model architecture with M$_0$, while its input description is replaced with the aspect.

\textit{Effect of Knowledge Integration Gate.}
Without KI Gate, M$_6$ obtains worse results than M$_0$.
This indicates that after DSG-Net, some aspect knowledge is further needed and KI Gate is efficient to re-enhance the final aspect representation with the needed knowledge.

\textit{Effect of Aspect-to-Context Attention.}
In M$_7$, the final aspect representation is used for prediction.
We can find that M$_7$ has limited performance degradation compared to M$_0$.
This proves that although previous modules can discover and extract clues for ASC, there are still important clues contained in non-aspect hidden states rather than final aspect representation.
Hence it is necessary to employ A2C Att to aggregate the aspect-related semantics in all hidden states into the final representation.

\textit{Effect of A2D Att and Self MHA in KaGR-MN Cell.}
%We first study the effect of A2D Att and Self MHA.
The significant performance decrease of M$_8$ shows that A2D Att is indispensable to dynamically summarize the specifically needed aspect knowledge from $\mathbf {M_D}$.
Without Self MHA, the integrated knowledge in aspect representation can not be contextualized and context memories cannot be updated.
As a result, M$_9$ performs much worse than M$_0$.
%On the other hand, the semantic consistent between the aspect representation and other context memories cannot be achieved, which harms the syntax graph modeling in DSG-Net.

%The large performance gaps also prove the effectiveness of learning the interaction of GCN and Relational MHA to capture mutual syntactic information.

\subsection{Investigation on Knowledge Gates} \label{sec: gate}
\begin{table*}[t]
\fontsize{8}{11}\selectfont
\centering
\caption{Results of different knowledge gate settings.}
\setlength{\tabcolsep}{2.5mm}{
\begin{tabular}{l|c|c|ccc}\toprule
\multirow{2}{*}{Variants} & \multirow{2}{*}{Gate 1} &\multirow{2}{*}{Gate 2} & Lap14 & Res14 & Res15 \\ \cline{4-6}
      & &                                                  & Acc   & Acc   & Acc   \\ \midrule
M$_0$          & AdaKI         & KI           &\textbf{81.87}     &\textbf{87.35 }     &\textbf{86.59} \\ \hline
M$_{10}$        & AdaKI     & AdaKI        & 81.50 ($\downarrow0.37$)  &  87.05 ($\downarrow0.30$) &  85.98 ($\downarrow0.61$)     \\
M$_{11}$        & KI            & KI     &  81.09 ($\downarrow0.78$)  &  86.73 ($\downarrow0.62$)  &   85.36 ($\downarrow1.23$) \\
M$_{12}$         & KI       & AdaKI        &  81.03 ($\downarrow0.84$) &  86.58 ($\downarrow0.77$)    &  85.24 ($\downarrow1.35$)  \\
\bottomrule
\end{tabular}}
\label{table: gate}
\end{table*}
KaGRMN-DSG has two different knowledge gates (AdaKI and KI) for knowledge integrating.
Here we empirically investigate these two knowledge gates by testing their four different settings.
The results are shown in Table \ref{table: gate}.
%We can find that respectively compared with M$_0$ and M$10$, M$_9$ and M$_11$ have slight decreases in performances.
We can find that M$_{10}$ and M$_{12}$ have slight decreases in performances when respectively compared with M$_0$ and M$_{11}$.
This is because KI Gate can preserve the knowledge in $\mathbf{r_k^\mathbf{T}}$ while AdaKI Gate may lose some knowledge when adapting to the semantic space of $\mathbf{\widetilde{R_a}}$.
M$_{11}$ and M$_{12}$ perform much worse than M$_0$ and M$_{10}$.
This is because the semantic space adaption of AdaKI Gate in KaGR-MN can maintain the semantics consistency of $\mathbf{r_a^{t*}}$ and $\mathbf{M_C^{t-1}}$, which is crucial for subsequent knowledge contextualizing.

\subsection{Impact of Time Step Number $\mathbf{T}$ }
\begin{figure*}[th]
 \centering
 \includegraphics[width =\textwidth]{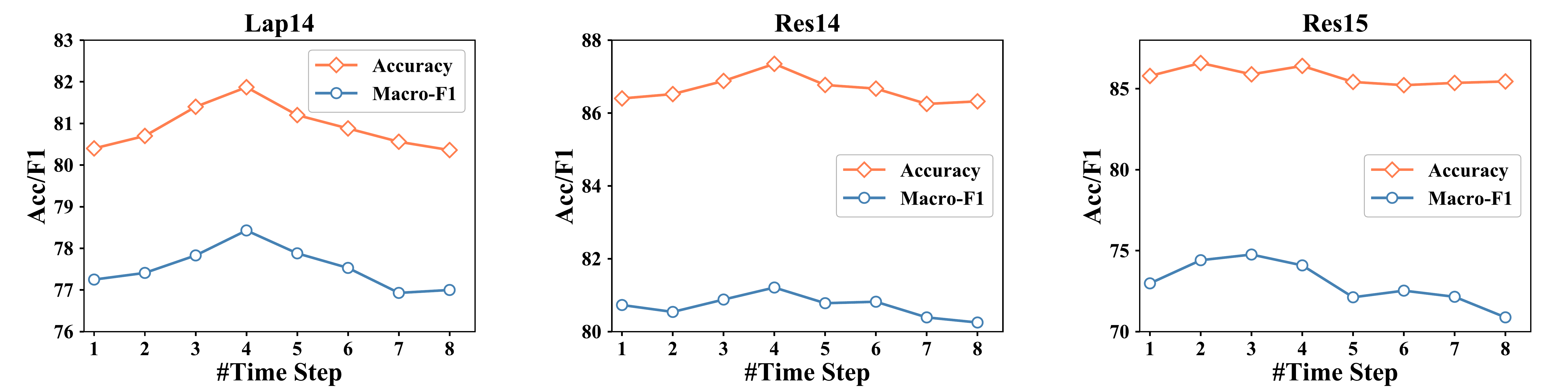}
 \caption{Impact of the time step number $\mathbf{T}$}
 \label{fig: t-effect}
\end{figure*}
We plot the performance trends of KaGRMN-DSG with increasing $\mathbf{T}$ on the three datasets, as presented in Fig. \ref{fig: t-effect}.
We can observe that the performances show a trend of increases at first and then decreases.
And the best result is obtained when $\mathbf{T}$ is 2 or 3 for Res15 and 4 for Lap14 and Res14.
This shows that appropriately increasing $\mathbf{T}$ can gradually improve the results, which is consistent with our expectation.
This can also prove the effectiveness of the recurrent manner of KaGR-MN.
However, too large $\mathbf{T}$ leads to inferior performances, which is also consistent with our expectation. One possible explanation is that too much knowledge integrated into the aspect representation and context memories will harm their original contextual information.
Another is that too many recurrent steps will lead to overfitting on training sets.
\subsection{Case Study} \label{sec: casestudy}
\begin{table*}[t]
\fontsize{8.9}{12}\selectfont
\centering
\caption{Cases demonstration. [\textbf{N}, \textbf{P}, \textbf{O}] denotes predicted sentiment distribution: [Negative, Positive, Neutral]. }
\setlength{\tabcolsep}{2mm}{
\begin{tabular}{l|l} \toprule
\textbf{Case} &  [\textbf{N}, \textbf{P}, \textbf{O}] \\ \hline
1. \tabincell{l}{$\mathbf{C}$: The \textbf{\textcolor{blue}{[Mountain Lion OS]}}$^\mathbf{A}$ is not hard to figure out if you are familiar with Microsoft Windows.\\ 
$\mathbf{D}$: OS X Mountain Lion is ... Apple Inc.'s desktop and server operating system ...}    &  \tabincell{l}{M$_0$: [0.0, \textbf{0.999}$^\checkmark$, 0.001] \\ M$_1$: [0.01, \textcolor{red}{0.49$^\times$}, 0.5]}        \\ \hline       
2. \tabincell{l}{$\mathbf{C}$: On start up it asks endless questions just so \textbf{\textcolor{blue}{[iTune]}}$^\mathbf{A}$ can sell you more of their products.\\ 
$\mathbf{D}$: iTunes is a media player, media library, Internet radio broadcaster, mobile device management utility ...}    &  \tabincell{l}{M$_0$: [\textbf{0.57}$^\checkmark$, 0.41, 0.02] \\ M$_1$: [0.03, \textcolor{red}{$0.67^\times$}, 0.30]}        \\ \hline
3. \tabincell{l}{$\mathbf{C}$: While the \textbf{\textcolor{blue}{[smoothies]}}$^\mathbf{A}$ are a little big for me, the fresh juices are the best i have ever had!\\ 
$\mathbf{D}$: A smoothie is a drink made from pureed raw fruit and/or vegetables, typically using a blender ...}    &  \tabincell{l}{M$_0$: [\textbf{0.62}$^\checkmark$, 0.0, 0.38] \\ M$_1$: [0.02, \textcolor{red}{$0.97^\times$}, 0.01]}        \\ \hline
%4. \tabincell{l}{$\mathbf{C}$: Did not enjoy the new Windows 8 and touchscreen functions.\\ 
%$\mathbf{D}$: Windows 8 is an operating system that was produced by Microsoft, released as part of ...}    &  \tabincell{l}{M$_0$: [\textbf{0.99}$^\checkmark$, 0.01, 0.0] \\ M$_2$: [\textbf{0.65}$^\checkmark$, 0.26, 0.09]}        \\ \hline \bottomrule
4. \tabincell{l}{$\mathbf{C}$: All the various Greek and Cypriot dishes are excellent, but the \textbf{\textcolor{blue}{[gyro]}}$^\mathbf{A}$ is the reason to come -- if you \\ don't eat one your trip was wasted.\qquad
$\mathbf{D}$: A gyro or gyros is a Greek dish made from meat cooked on a ...}    &  \tabincell{l}{M$_0$: [0.02, \textbf{0.98}$^\checkmark$, 0.0] \\ M$_1$: [\textcolor{red}{0.88$^\times$}, 0.11, 0.01]} \\ \bottomrule
\end{tabular}}
\label{table: case}
\end{table*}
We show some cases in Table \ref{table: case}. Note that the only difference between KaGRMN-DSG (M$_0$) and M$_1$ is that the input $\mathbf{D}$ in M$_1$ is replaced with $\mathbf{A}$.
We can observe that M$_0$ can accurately predict the correct labels in all cases, while M$_1$ fails all cases although its overall performance is promising (as shown in Table \ref{table: ablation})
%This is because M$_0$ can effectively integrate the needed aspect knowledge into the aspect representation and context
%memories. The integrated knowledge can provide appropriate aspect semantics and help the model capture important clues for ASC.

Without leveraging aspect knowledge, the aspect representation and semantics derived by M$_1$ are inadequate. 
%basic processing unit of the encoder is a token (or subword) rather than an entity. 
As shown in Table \ref{table: entity}, BERT cannot capture the exact meanings and properties of \textit{Mountain Lion OS} and \textit{iTune}, although it is one of the strongest language models. 
In Case 1, M$_1$ regards \textit{Mountain Lion OS} as `lion' which is `dangerous'. Then considering `not hard', M$_1$ is confused on P and O.
In contrast, leveraging aspect knowledge, M$_0$ captures the exact meaning: an operating system.
Then considering the aspect-related semantics (`not hard'), M$_0$ correctly predicts P.
In Case 2, the aspect sentiment expression is a little obscure as there are no explicit sentiment trigger words (e.g. delicious, good, expensive).
Even if M$_1$ captures aspect-related context semantics, it fails due to the lack of property information of \textit{iTune}.
Thanks to the integrated aspect knowledge, M$_0$ is aware that \textit{iTune} is primarily used for media playing rather than selling products, thus correctly predicts N.
%Looking into case 2, the exact semantics of \textit{Mountain Lion OS} is far from the semantics combination of each token.

Looking into Case 3 and Case 4, we can find that due to the lack of aspect knowledge, M$_1$ is prone to be affected by some misleading sentiment trigger words: `best' in case 3 and `but' in case 4.
The reason why M$_0$ wins M$_1$ is that M$_0$ can combine the aspect knowledge and the aspect-related semantics together to capture the correct clues for ASC.
\begin{table}[t]
\fontsize{8}{10}\selectfont
\centering
\caption{Misunderstanding from BERT presented by semantic cosine similarity ($S$). $v$ is the average of entity's hidden states.
 $a_i$ denotes the $\mathbf{A}$ in case $i$.}
\setlength{\tabcolsep}{2mm}{
\begin{tabular}{c|c||c|c}\toprule
{Entity ($e$)} & $S(v_e, v_{a_1})$ &{Entity ($e$)} & $S(v_e, v_{a_2})$          \\ \midrule
lion     & 0.8516   &media player       &0.4720    \\
mountain           &0.7997  &   radio broadcaster& 0.5887        \\ 
operating system & 0.6826       & software & 0.7051        \\
dangerous animal & 0.8272 & utility & 0.6982\\
          \bottomrule
\end{tabular}}
\label{table: entity}
\end{table}

\subsection{Computation Time Analysis}
\begin{table}[t]
\fontsize{8}{10}\selectfont
\centering
\caption{Comparison of training time and inference time (per sample) as well as the avg F1 on the three datasets.}
\setlength{\tabcolsep}{2mm}{
\begin{tabular}{c|c|c|c}\toprule
Models & Training Time$\downarrow$ & Inference Time$\downarrow$ &Avg F1$\uparrow$ \\ \midrule
BERT-SPC & \textbf{0.007309}s &\textbf{0.002219}s   &    73.59\%    \\
BERT+T-GCN & 0.033835s & 0.003350s  &   76.22\%    \\ \hline
KaGRMN-DSG  &0.015333s & 0.004208s  &   \textbf{78.91}\%   \\ \bottomrule
\end{tabular}}
\label{table: time}
\end{table}
The comparison of time costing and avg F1 of BERT-SPC, BERT+T-GCN and our KaGRMN-DSG model is shown in Table \ref{table: time}.
We can find that although our model demands more training time and inference time than BERT-SPC, it overpasses BERT-SPC on avg F1 by a large margin (6.3\%).
As for BERT+T-GCN, which is the best-performing baseline, although it costs lightly less inference time than our KaGRMN-DSG, it costs much more time for training, and more importantly, its performance is significantly inferior to us.
Additionally, since \textit{Local Syntactic Information Modeling} and \textit{Global Relational Information Modeling} both take the output of KaGRMN as input, they can be parallelized theoretically, so the training time and inference time of our KaGRMN-DSG model can be further reduced in practice.
In a word, our model may cost more time for training and inference than some baseline models, but it is worthy considering the significant performance improvement.

%% file: conclusion.tex
% !TEX root = bare_jrnl.tex
\section{Conclusion}
In this paper, we point out the two challenges encountering existing ASC models and we therefore propose a novel KaGRMN-DSG model to end-to-end embed and leverage aspect knowledge, then capture sufficient syntactic information by marrying both kinds of syntactic information.
In our model, the integrated beneficial aspect knowledge and sufficient syntactic information can effectively cooperate, yielding new state-of-the-art results.

Future directions include exploring the visual knowledge of aspects, as well as designing deeper and more sufficient dual syntactic interaction to let the two kinds of syntactic information interact with each other in their respective modeling processes.